\pdfoutput=1
\documentclass[11pt]{article}

\usepackage[]{acl}
\usepackage{hhline}
\usepackage{framed}
\usepackage{mathtools}
\usepackage{times}
\usepackage{latexsym}

\usepackage{multirow}
\usepackage{url}
\usepackage{amsmath}
\usepackage{bbm}
\usepackage{array}
\usepackage{utfsym}
\usepackage{subfigure}
\usepackage{balance}
\usepackage[misc]{ifsym}
\usepackage{graphicx}
\usepackage{enumitem}
\usepackage{makecell}
\usepackage{booktabs}
\usepackage{amssymb}
\usepackage{color,xcolor}
\definecolor{darkgreen}{HTML}{3CB371}
\definecolor{darkred}{HTML}{7B7B7B}
\usepackage{cases}
\usepackage{colortbl} 
\usepackage{diagbox}
\definecolor{ggreen}{HTML}{d4e7cf}
\definecolor{bblue}{HTML}{00bfff}
\definecolor{ppink}{HTML}{ff69b4}

\usepackage[T1]{fontenc}
\usepackage[utf8]{inputenc}

\usepackage{microtype}

\title{Cross-Lingual Knowledge Editing in Large Language Models}

\author{Jiaan Wang\textsuperscript{$\spadesuit$}, \ Yunlong Liang\textsuperscript{$\heartsuit$}, \ Zengkui Sun\textsuperscript{$\diamondsuit$}, \ Yuxuan Cao\textsuperscript{$\clubsuit$}, \\
\bf {Jiarong Xu\textsuperscript{$\spadesuit$}\thanks{ \ \ Corresponding author.} \ and Fandong Meng\textsuperscript{$\heartsuit$}} \\
\\
\textsuperscript{$\spadesuit$}Fudan University \qquad \textsuperscript{$\heartsuit$}Pattern Recognition Center, WeChat AI, Tencent Inc, China \\
\textsuperscript{$\diamondsuit$}Beijing Jiaotong University \qquad \textsuperscript{$\clubsuit$}Zhejiang University  \\
\small \texttt{jawang.nlp@gmail.com} \quad \texttt{\{yunlonliang,fandongmeng\}@tencent.com} \\
\small \texttt{zengksun@bjtu.edu.cn} \quad \texttt{caoyx@zju.edu.cn} \quad \texttt{jiarongxu@fudan.edu.cn}\\
}

\begin{document}
\maketitle
\begin{abstract}

Knowledge editing aims to change language models' performance on several special cases (\emph{i.e.}, editing scope) by infusing the corresponding expected knowledge into them. With the recent advancements in large language models (LLMs), knowledge editing has been shown as a promising technique to adapt LLMs to new knowledge without retraining from scratch. However, most of the previous studies neglect the multi-lingual nature of some main-stream LLMs (\emph{e.g.}, LLaMA, ChatGPT and GPT-4), and typically focus on monolingual scenarios, where LLMs are edited and evaluated in the same language.
As a result, it is still unknown the effect of source language editing on a different target language. In this paper, we aim to figure out this cross-lingual effect in knowledge editing. Specifically, we first collect a large-scale cross-lingual synthetic dataset by translating ZsRE from English to Chinese.
Then, we conduct English editing on various knowledge editing methods covering different paradigms, and evaluate their performance in Chinese, and vice versa.
To give deeper analyses of the cross-lingual effect, the evaluation includes four aspects, \emph{i.e.}, reliability, generality, locality and portability.
Furthermore, we analyze the inconsistent behaviors of the edited models and discuss their specific challenges.\footnote{Data and codes are available at \url{https://github.com/krystalan/Bi_ZsRE}}

\end{abstract}

\section{Introduction}

\begin{figure}[t]
\centerline{\includegraphics[width=0.48\textwidth]{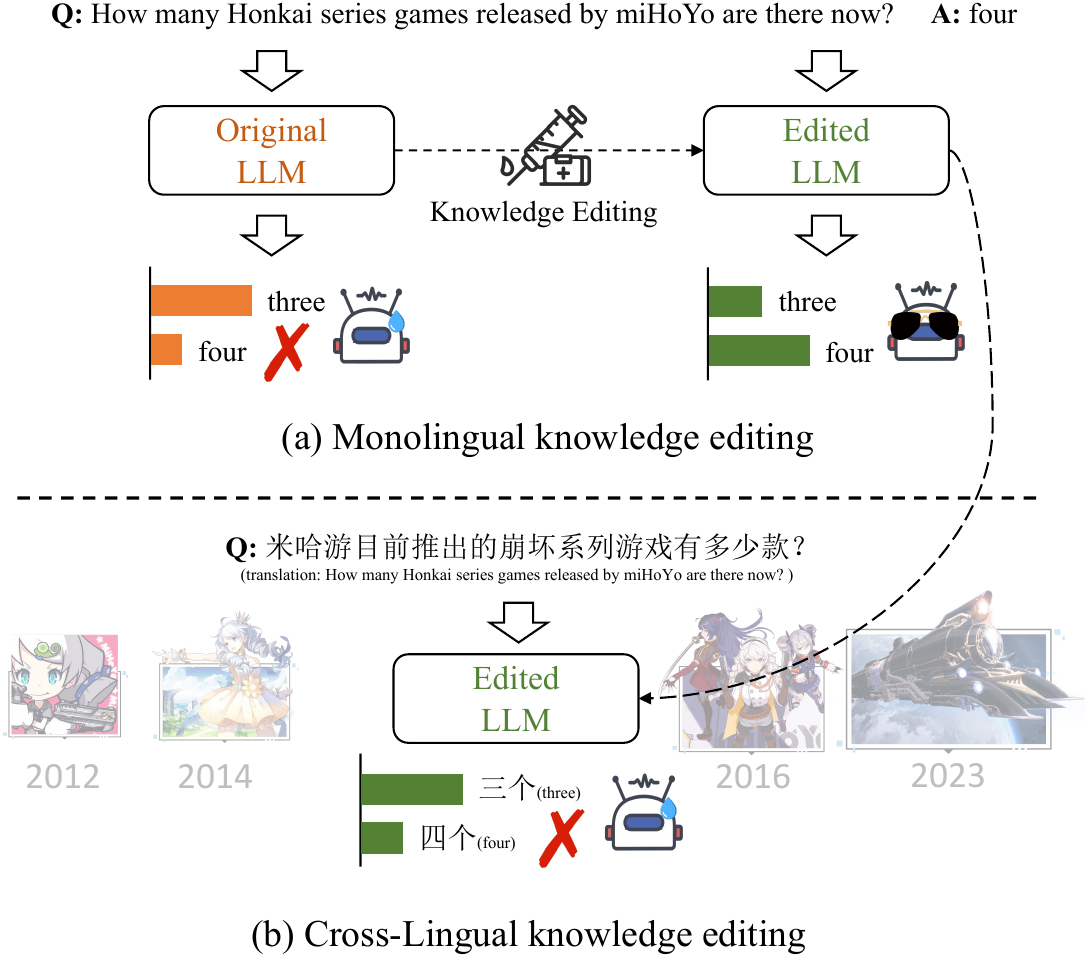}}
\caption{Illustration of (a) monolingual knowledge editing, where the model is edited and verified in the same language; and (b) cross-lingual knowledge editing, where the model is edited and verified in different languages.}
\label{fig:intro}
% \vspace{-0.5cm}
\end{figure}

The goal of knowledge editing is to adjust language models' behaviors within an expected scope (\emph{i.e.}, editing scope) and retain out-of-scope model performance ideally~\cite{Yao2023EditingLL}.
Along with the dynamic changes in the world, knowledge editing could help models forget outdated knowledge and adapt to the new counterpart without retraining from scratch.
As the example shown in Figure~\ref{fig:intro} (a), the number of Honkai-series games increases to four after the release of \emph{Honkai: Star Rail} (on April 26, 2023). However, if we ask a model that has been trained before the date, the model might only know three Honkai-series games.
In such a situation, knowledge editing could help the model efficiently update this new knowledge, and give the right answer after editing.

Despite many efforts devoted to this research field~\cite{de-cao-etal-2021-editing,mitchell2022memory,dong-etal-2022-calibrating,dai-etal-2022-knowledge,meng2022locating,mitchell2022fast,huang2023transformer,meng2023mass,zheng2023can,wang2024detoxifying,wang2024wise,ma2024neighboring,zhang2024comprehensive,zhang-etal-2024-knowledge,wang2024editing}, current knowledge editing studies typically focus on monolingual scenarios, where language models are edited and evaluated within the same language, c.f., Figure~\ref{fig:intro} (a).
Meanwhile, the rapid advancements in large language models (LLMs) have led to the widespread adoption of multi-lingual settings, allowing language modeling ability can be shared across different languages~\cite{zhao2023survey,wang-etal-2023-zero,wang-etal-2022-survey,liang-etal-2023-d2tv,liang-etal-2023-summary,wang-etal-2023-towards-unifying,wang-etal-2022-clidsum}.
For example, LLMs such as LLaMA~\cite{touvron2023llama}, ChatGPT~\cite{ChatGPT}, and GPT-4~\cite{OpenAI2023GPT4TR} are designed to operate under multilingual setting.
Under this background, the performance of a source-language edited model on other languages is still unknown.
As shown in Figure~\ref{fig:intro} (b), a research question (RQ) arises, \emph{when we utilize source-language samples to edit a multi-lingual LLM, can the model reflect consistent behaviors when faced with a different target language?}

To answer the RQ, in this paper, we explore knowledge editing in cross-lingual scenarios, and study the effects of source-language editing on a different target language. Specifically, we automatically translate the knowledge editing data from English to Chinese via cutting-edge LLMs (\emph{i.e.}, ChatGPT and GPT-4).
After carefully comparing existing datasets, we finally choose ZsRE~\cite{levy-etal-2017-zero} which is originally a question answering (QA) dataset and is further widely used in knowledge editing~\cite{de-cao-etal-2021-editing,meng2022locating,mitchell2022fast}.
More recently, \citet{Yao2023EditingLL} collect a number of QA pairs that need deep reasoning based on ZsRE, and the data could be used to evaluate the portability of knowledge editing methods beyond simple paraphrasing.
Therefore, we also translate these QA pairs to give a deeper understanding of cross-lingual knowledge editing performance.
The translated data together with the original ones is denoted as Bi-ZsRE.
Then, we conduct English/Chinese editing on several open-sourced multi-lingual LLMs (LLaMA, LLaMA2, Baichuan and Baichuan2), and evaluate their behaviors in Chinese/English in terms of reliability, generality, locality and portability.
Our experiments involve seven knowledge editing methods covering three main-stream paradigms pointed by~\citet{Yao2023EditingLL}, \emph{i.e.}, memory-based, meta-learning and locate-then-edit methods.
The experimental results reveal that (1) the language modeling gaps across different languages influence the efficiency of knowledge editing; (2) it is still hard for existing knowledge editing methods to transfer the edited knowledge from one language to another in a multi-lingual LLM; (3) when editing LLMs in a language, the model performance on the irrelevant examples in other languages could also be influenced, resulting in low locality.
This presents a significant challenge for multi-lingual LLMs in maintaining consistent behaviors across different languages.

Our main contributions are concluded as follows:
\begin{itemize}[leftmargin=*,topsep=0pt]
\setlength{\itemsep}{0pt}
\setlength{\parsep}{0pt}
\setlength{\parskip}{0pt}
\item To our knowledge, we are the first to explore the cross-lingual effect of knowledge editing in LLMs. We achieved this by translating the ZsRE dataset and studying the cross-lingual effect from English (Chinese) to Chinese (English). 
\item We conduct experiments on various knowledge editing methods and multi-lingual LLMs. Our results indicate that it remains challenging for multi-lingual LLMs to generalize the edited knowledge to other languages.
\item In-depth analysis of the inconsistent behaviors exhibited by the edited models and their specific challenges provide us with a deeper understanding of the cross-lingual effect in knowledge editing.
\end{itemize}

\section{Related Work}

\noindent \textbf{Knowledge Editing Methods.} The goal of knowledge editing is to alter the behavior of LLMs within an expected scope (\emph{i.e.}, editing scope) without negatively impacting performance out of the scope.
According to a comprehensive survey on knowledge editing~\cite{Yao2023EditingLL}, there are three main-stream knowledge editing paradigms: (1) \emph{Memory-based methods} keep the original model parameters unchanged while employing another model to influence the model’s behaviors. SERAC~\cite{mitchell2022memory} utilizes a scope classifier to evaluate whether new input is close to the stored editing examples, and further influences the model behaviors based on the retrieved editing examples.
T-Patcher~\cite{huang2023transformer} and CaliNET~\cite{dong-etal-2022-calibrating} add extra trainable parameters into the FFN layers of LLMs to edit model performance.
IKE~\cite{zheng2023can} uses context-edit facts to guide the model in generating edited facts.
(2) \emph{Meta-learning methods} employ a hyper network to learn the weight updates of LLMs to edit the models. KE~\cite{de-cao-etal-2021-editing} makes use of LSTM networks to predict the weight update for each new input.
MEND~\cite{mitchell2022fast} transforms the gradient of fine-tuned language models by employing a low-rank decomposition of gradients.
(3) \emph{Locate-then-edit methods} first identify parameters corresponding to specific knowledge and then update these parameters.
Among them, KN~\cite{dai-etal-2022-knowledge} specifies a key-value pair in the FFN matrix that embodies the knowledge and then proceeds to update the corresponding parameters.
ROME~\cite{meng2022locating} leverages causal mediation analysis to locate the edit area, and update the whole parameters in the FFN matrix.
MEMIT~\cite{meng2023mass} directly updates LLMs with many memories, and thus, facilitating thousands of edits to be executed simultaneously.

More recently, \citet{xu-etal-2023-language-anisotropic} introduce the cross-lingual model editing task and design a language anisotropic editing method. However, the proposed method is applied to mBERT~\cite{devlin-etal-2019-bert} (a classical pre-trained multi-lingual NLU model), making the cross-lingual effect still unknown in generative LLMs. 

\begin{table*}[t]
\centering
\resizebox{0.90\textwidth}{!}
{
\begin{tabular}{lccccccccc}
\toprule[1pt]
\multirow{2}{*}{Splitting}  & \multirow{2}{*}{Lang.} & \multirow{2}{*}{\# Example} & \multirow{2}{*}{Question} & Rephrased  & \multirow{2}{*}{Answer}  & Locality & Locality & Portability & Portability \\
  &   &    &   &     Question     &   & Question & Answer   & Question    & Answer      \\ \midrule[1pt]
\multirow{2}{*}{Training} & En   & 10,000     & 11.28   &  11.25   & 2.85     & 15.25    & 5.61     & -           & -           \\
& Zh     & 10,000   & 10.86    &   10.95      & 4.36     & 14.71    & 6.77     & -           & -           \\
\multirow{2}{*}{Validation} & En   & 3,000     & 11.19      &   11.20                  & 2.79      & 15.39    & 5.50      & -           & -           \\
  & Zh     & 3,000      & 10.94     & 11.01      &  4.37     & 14.66    & 6.55     & -           & -           \\
\multirow{2}{*}{Test}       & En      & 1,037       & 11.43     &   11.49         & 3.11     & 15.31    & 5.62     & 18.02       & 4.54        \\
  & Zh      & 1,037       & 11.48    & 11.60     & 4.69      & 11.56    & 6.77     & 16.48       & 5.88        \\ \bottomrule[1pt]
\end{tabular}
}
\caption{Statistics of Bi-ZsRE (Lang.: language; En: English; Zh: Chinese). ``\emph{\# Example}'' indicates the number of samples in each subset. All decimals denote the average length (token-level) of different aspects in each subset.}
\label{table:statistics}
\end{table*}

\vspace{0.5ex}
\noindent \textbf{Knowledge Editing Datasets.} ZsRE~\cite{levy-etal-2017-zero} is a question answering dataset whose queries require models to answer the questions based on the information within the queries.
\textsc{CounterFact}~\cite{meng2022locating} evaluates whether the edited model can provide counterfactual answers when asked about the corresponding factual knowledge.
\textsc{MQuAKE}~\cite{zhong2023mquake} aims to assess whether edited models correctly answer questions where the answer needs reasoning based on the edited facts.
\citet{Cheng2023CanWE} propose MMEdit, a multi-modal knowledge editing benchmark dataset.
PersonalityEdit~\cite{Mao2023EditingPF} is proposed to edit personality traits for LLMs.
\citet{li2023unveiling} propose ConflictEdit and RoundEdit to investigate the potential pitfalls associated with knowledge editing for LLMs.
Eva-KELLM~\cite{wu2023eva} evaluates the edited model from reasoning with the altered knowledge and cross-lingual transfer.
Though Eva-KELLM provides a subset for cross-lingual knowledge editing, the data has not yet been made public.\footnote{October 15, 2023}
Besides, this work does not conduct experiments on any knowledge editing methods, leaving the cross-lingual effect still not known in the knowledge editing research field.

\section{Bi-ZsRE}

In this section, we first discuss the details of data collection, including data sources, translation process as well as quality control (\S~\ref{subsec:3.1}).
Then, we give the data statistics of Bi-ZsRE (\S~\ref{subsec:3.2}), and finally provide the task overview of cross-lingual knowledge editing (\S~\ref{subsec:3.3}).

\subsection{Data Collection}
\label{subsec:3.1}

\noindent \textbf{Data Sources.}
ZsRE~\cite{levy-etal-2017-zero} is a Question Answering (QA) dataset whose queries require models to answer the questions based on the information within the queries. 
Following previous data settings~\cite{Yao2023EditingLL,wang2023easyedit}, it contains 163,196 training samples and 19,086 validation samples.
Each sample involves a question and a corresponding answer for editing LLMs. To evaluate the generality of edited models, a rephrased question is also provided.
Besides, each sample also associates with an unrelated QA pair (selected from the NQ dataset~\cite{kwiatkowski-etal-2019-natural}) to evaluate the locality.
Recently, \citet{Yao2023EditingLL} provide a test set with 1,037 samples for a more comprehensive evaluation of knowledge editing, where each test sample additionally contains a QA pair to assess LLMs’ portability to reason based on the edited fact.
To control the cost of translation, we randomly selected 10,000 training samples and 3,000 validation samples, which together with all test samples are further translated.

\vspace{0.5ex}
\noindent \textbf{Translation Process.} We use \texttt{gpt-3.5-turbo} and \texttt{gpt-4} to translate the above knowledge editing data from English to Chinese. In particular, considering the trade-off between quality and cost, training samples and validation samples are translated by \texttt{gpt-3.5-turbo}, while test samples are translated by \texttt{gpt-4}.
The translation is conducted based on the OpenAI's official APIs\footnote{\url{https://platform.openai.com/docs/api-reference/chat/object}} with zero temperature.
The used translation prompt is shown as follows:
\begin{framed}
\small
\noindent Please translate the following JSON data from English to Chinese and keep the format unchanged:

\noindent \textcolor[RGB]{101,42,150}{\texttt{[JSON data]}}
\end{framed}
\noindent where each sample is organized in JSON format and further translated at the sample level.

\vspace{0.5ex}
\noindent \textbf{Quality Control.} To further ensure the translation quality of the test samples, we also employ three translators to correct the translations of \texttt{gpt-4}.
All translators are native Chinese and are fluent in English.
As a result, there are about 6.0\% of samples are corrected while the remaining are unchanged.
All corrected samples are further checked by a data expert who has rich experience in translation annotations.
Finally, all translated data and original data are denoted as Bi-ZsRE.

\subsection{Data Statistics}
\label{subsec:3.2}

Table~\ref{table:statistics} lists the data statistics of Bi-ZsRE, covering two languages, English (En) and Chinese (Zh), across three subsets.
For English samples, the average question lengths are 11.28, 11.19, and 11.43 tokens in the training, validation, and test subsets, respectively, while the counterparts in Chinese are 10.86, 10.94, and 11.48.
Besides, the average length of portability questions is longer than that of original questions, rephrased questions or locality questions, thus portability questions may involve more intricate reasoning based on the edited knowledge. To correctly answer the portability questions, the edited model should absorb the knowledge rather than simply memorize the word replacement.

\subsection{Task Overview}
\label{subsec:3.3}

\noindent \textbf{Knowledge Editing.} Given a language model $p_\theta$ and an edit descriptor $\langle x_e, y_e \rangle$, the goal of knowledge editing is to create an edited model $p'_\theta$ satisfy the following requirements:
\begin{equation}
p'_\theta (x) = 
\begin{cases}
y_e & x \in \mathcal{X}_e \\
p_\theta (x) & x \notin \mathcal{X}_e 
\end{cases}
\end{equation}
where $\mathcal{X}_e$ denotes a broad set of inputs with the same semantics as $x_e$.
The edited model should also satisfy the following four properties:
(1) \emph{Reliability} measures the average accuracy on the edit case. When receiving $x_e$ as input, the edited model $p'_\theta$ should output $y_e$.
(2) \emph{Generality} evaluates the average accuracy on the equivalent cases as the edit case. For instance, when receiving a rephrased text of $x_e$, the edited model $p'_\theta$ is also expected to output $y_e$.
(3) \emph{Locality} assesses the accuracy of the edited model on the irrelevant samples.
When the input $x$ is out of the edit scope $\mathcal{X}_e$, $p'_\theta (x)$ should be the same as $p_\theta (x)$ ideally.
(4) \emph{Portability} measures the robust generalization of the edited model via a portability question that needs reasoning based on the edited knowledge. When receiving the portability question as input, the edited model $p'_\theta$ is expected to output the golden answer to demonstrate the model indeed learns the knowledge rather than memorizing superficial changes in wording.

\begin{table*}[t]
\centering
\resizebox{0.95\textwidth}{!}
{
\begin{tabular}{llccccccc}
\toprule[1pt]
\multicolumn{1}{c}{Editing} & \multicolumn{1}{c}{\multirow{2}{*}{Method}}&  \multirow{2}{*}{Reliability} & \multicolumn{3}{c}{English Evaluation} & \multicolumn{3}{c}{Chinese Evaluation}    \\
\multicolumn{1}{c}{Language}    &                    &                              & Generality               & Locality              & Portability            & Generality               & Locality              & Portability             \\ \midrule[1pt]
\multicolumn{9}{c}{\texttt{Chinese-LLaMA-Plus-7B}} \\ \midrule[1pt]
\multicolumn{1}{c}{\multirow{7}{*}{English}} & FT                                       & 20.46 / \textcolor{darkred}{00.77}                & 18.36 / \textcolor{darkred}{00.19}   & \textcolor{darkgreen}{87.49} / 70.11  & \textcolor{darkred}{06.30} / \textcolor{darkred}{00.00}  & 22.08 / \textcolor{darkred}{00.10}    & 79.52 / 47.44     & 24.63 / \textcolor{darkred}{00.00}  \\
& SERAC                                  & 73.84 / 56.03                & 50.86 / 27.10  & \textcolor{darkgreen}{100.0} / \textcolor{darkgreen}{100.0}  & \textcolor{darkred}{06.52} / \textcolor{darkred}{00.00}  & 19.26 / \textcolor{darkred}{00.29}    & \textcolor{darkgreen}{99.97} / \textcolor{darkgreen}{99.90}    & 15.63 / \textcolor{darkred}{00.00}  \\
& IKE                                     & \textcolor{darkgreen}{99.90} / \textcolor{darkgreen}{99.90}   & \textcolor{darkgreen}{99.24} / \textcolor{darkgreen}{98.36}  & 62.79 / 36.26  & 50.86 / 17.84   & \textcolor{darkgreen}{92.95} / 69.72      & 36.16 / \textcolor{darkred}{06.75}   & 33.84 / \textcolor{darkred}{04.24}  \\
& MEND                                   & 37.57 / \textcolor{darkred}{02.22}                & 33.24 / \textcolor{darkred}{01.35}   & \textcolor{darkgreen}{88.96} / 74.25  & \textcolor{darkred}{06.56} / \textcolor{darkred}{00.10}    & 17.08 / \textcolor{darkred}{00.00}   & \textcolor{darkgreen}{91.75} / 75.89   & 16.91 / \textcolor{darkred}{00.00}  \\
& KN                                     & \textcolor{darkred}{04.63} / \textcolor{darkred}{00.00}                & \textcolor{darkred}{04.54} / \textcolor{darkred}{00.00} & 42.25 / 29.12  & \textcolor{darkred}{03.53} / \textcolor{darkred}{00.00}   & \textcolor{darkred}{06.66} / \textcolor{darkred}{00.00}      & 36.75 / 19.67   & \textcolor{darkred}{08.46} / \textcolor{darkred}{00.00}  \\
& ROME                                    & \textcolor{darkgreen}{98.98} / \textcolor{darkgreen}{97.20}                & \textcolor{darkgreen}{94.58} / \textcolor{darkgreen}{87.85}   & \textcolor{darkgreen}{92.49} / \textcolor{darkgreen}{81.49}  & \textcolor{darkred}{08.48} / \textcolor{darkred}{00.00} & 26.65 / \textcolor{darkred}{06.75}    & \textcolor{darkgreen}{89.08} / 67.60     & 17.07 / \textcolor{darkred}{00.00}  \\
& MEMIT                                  & \textcolor{darkgreen}{96.19} / \textcolor{darkgreen}{92.48}                & \textcolor{darkgreen}{90.66} / \textcolor{darkgreen}{81.97}  & \textcolor{darkgreen}{98.31} / \textcolor{darkgreen}{94.70}  & \textcolor{darkred}{08.27} / \textcolor{darkred}{00.00}  & 28.26 / \textcolor{darkred}{06.75}    & \textcolor{darkgreen}{97.31} / \textcolor{darkgreen}{91.51}   & 17.96 / \textcolor{darkred}{00.00}  \\ \hline
\multicolumn{1}{c}{\multirow{7}{*}{Chinese}} & FT                                     & \textcolor{darkred}{09.54} / \textcolor{darkred}{00.19}                & 12.38 / \textcolor{darkred}{00.00}  & \textcolor{darkgreen}{87.81} / 73.29  & \textcolor{darkred}{06.77} / \textcolor{darkred}{00.10} & 35.91 / \textcolor{darkred}{00.96}    & 57.78 / 15.24   & 21.09 / \textcolor{darkred}{00.19}  \\
& SERAC                                 & 27.05 / 12.83                & 14.67 / \textcolor{darkred}{00.00}   & \textcolor{darkgreen}{100.0} / \textcolor{darkgreen}{100.0} & \textcolor{darkred}{06.56} / \textcolor{darkred}{00.00}  & 67.41 / 37.32  & \textcolor{darkgreen}{94.42} / \textcolor{darkgreen}{85.82}  & 20.65 / \textcolor{darkred}{00.00}  \\
& IKE                                    & \textcolor{darkgreen}{99.90} / \textcolor{darkgreen}{99.71}                & \textcolor{darkgreen}{85.39} / 77.24   & 64.14 / 37.32  & 40.07 / \textcolor{darkred}{05.01}  & \textcolor{darkgreen}{97.31} / \textcolor{darkgreen}{95.37}     & 52.46 / 17.36   & 38.39 / \textcolor{darkred}{07.52}  \\
& MEND                                   & 15.47 / \textcolor{darkred}{00.48}                & 14.39 / \textcolor{darkred}{00.00}  & \textcolor{darkgreen}{89.19} / 73.87  & \textcolor{darkred}{06.72} / \textcolor{darkred}{00.10}  & 44.32 / \textcolor{darkred}{00.68}    & 78.17 / 46.00   & 22.94 / \textcolor{darkred}{00.19}  \\
& KN                                      & \textcolor{darkred}{03.09} / \textcolor{darkred}{00.00}                & \textcolor{darkred}{04.74} / \textcolor{darkred}{00.00}   & 29.82 / 16.39  & \textcolor{darkred}{02.87} / \textcolor{darkred}{00.00}  & \textcolor{darkred}{05.08} / \textcolor{darkred}{00.00}     & 20.08 / \textcolor{darkred}{08.78}   & \textcolor{darkred}{05.56} / \textcolor{darkred}{00.00}  \\
& ROME                                    & 36.63 / 20.15                & 24.24 /  \textcolor{darkred}{08.87}   & \textcolor{darkgreen}{89.21} / 74.73  & \textcolor{darkred}{06.52} / \textcolor{darkred}{00.00}  & \textcolor{darkgreen}{81.83} / 39.92    & \textcolor{darkgreen}{86.44} / 63.74   & 21.33 / \textcolor{darkred}{00.10}  \\
& MEMIT                                  & 35.54 / 19.19                & 22.88 / \textcolor{darkred}{08.97}   & \textcolor{darkgreen}{98.13} / \textcolor{darkgreen}{94.12}  & \textcolor{darkred}{06.88} / \textcolor{darkred}{00.00} & \textcolor{darkgreen}{81.11} / 39.34    & \textcolor{darkgreen}{95.84} / \textcolor{darkgreen}{86.98}    & 23.29 / \textcolor{darkred}{00.00}  \\ \midrule[1pt]
\multicolumn{9}{c}{\texttt{Chinese-LLaMA-2-7B}} \\ \midrule[1pt]
\multicolumn{1}{c}{\multirow{7}{*}{English}} & FT                                      & 36.62 / \textcolor{darkred}{05.98}                & 35.01 / \textcolor{darkred}{07.52}   & \textcolor{darkgreen}{81.90} / 55.06 & \textcolor{darkred}{07.33} / \textcolor{darkred}{00.00}  & 20.24 / \textcolor{darkred}{00.10}   & 72.95 / 32.11  & 17.91 / \textcolor{darkred}{00.00}  \\
& SERAC                                       & \textcolor{darkgreen}{98.78} / \textcolor{darkgreen}{97.01}                & \textcolor{darkgreen}{89.62} / \textcolor{darkgreen}{82.64}     & \textcolor{darkgreen}{100.0} / \textcolor{darkgreen}{100.0} & \textcolor{darkred}{08.75} / \textcolor{darkred}{00.00}  & 21.92 /  \textcolor{darkred}{02.60} & \textcolor{darkgreen}{97.67} / \textcolor{darkgreen}{93.44}  & 17.30 / \textcolor{darkred}{00.00}  \\
& IKE                                          & \textcolor{darkgreen}{100.0} / \textcolor{darkgreen}{100.0}                & \textcolor{darkgreen}{99.69} / \textcolor{darkgreen}{99.32}   & 56.35 / 30.76 & 45.72 / 11.76  & \textcolor{darkgreen}{92.28} / 77.72  & 41.59 / \textcolor{darkred}{04.63}  & 37.04 / \textcolor{darkred}{04.82}  \\
& MEND                                         & 56.57 / \textcolor{darkred}{00.00}                & 49.33 / \textcolor{darkred}{00.00}    & \textcolor{darkgreen}{95.46} / \textcolor{darkgreen}{86.79}  & \textcolor{darkred}{07.62} / \textcolor{darkred}{00.00} & 20.66 / \textcolor{darkred}{00.00}   & \textcolor{darkgreen}{95.25} / \textcolor{darkgreen}{86.21}  & 17.34 / \textcolor{darkred}{00.00}  \\
& KN                                          & 10.94 / \textcolor{darkred}{00.00}                & 10.96 / \textcolor{darkred}{00.00} & 49.28 / \textcolor{darkred}{06.85}  & \textcolor{darkred}{05.75} / \textcolor{darkred}{00.00}  & 12.30 / \textcolor{darkred}{00.00}   & 43.65 / \textcolor{darkred}{09.35}  & 14.39 / \textcolor{darkred}{00.00}  \\
& ROME                                    & 77.65 / 67.98                & 72.27 / 55.06    & \textcolor{darkgreen}{93.67} / \textcolor{darkgreen}{81.58} & \textcolor{darkred}{07.48} / \textcolor{darkred}{00.10} & 23.27 / \textcolor{darkred}{03.28}  & \textcolor{darkgreen}{95.55} / \textcolor{darkgreen}{84.96}   & 17.88 / \textcolor{darkred}{00.00}  \\
& MEMIT                                    & \textcolor{darkgreen}{83.01} / 74.64                & 77.63 / 61.43    & \textcolor{darkgreen}{98.45} / \textcolor{darkgreen}{95.37} & \textcolor{darkred}{08.08} / \textcolor{darkred}{00.10} & 23.91 / \textcolor{darkred}{03.95}   & \textcolor{darkgreen}{98.13} / \textcolor{darkgreen}{93.54}   & 17.22 / \textcolor{darkred}{00.00}  \\ \hline
\multicolumn{1}{c}{\multirow{7}{*}{Chinese}} & FT                                      & 13.03 / \textcolor{darkred}{01.16}                & 16.30 / \textcolor{darkred}{01.06}  & 76.68 / 48.02 & \textcolor{darkred}{07.07} / \textcolor{darkred}{00.00}  & 36.01 / \textcolor{darkred}{00.77}   & 59.70 / 16.59  & 19.25 / \textcolor{darkred}{00.00}  \\
& SERAC         & 26.76 / 20.44                & 19.87 / \textcolor{darkred}{02.31}  & \textcolor{darkgreen}{100.0} / \textcolor{darkgreen}{100.0}  & \textcolor{darkred}{08.14} / \textcolor{darkred}{00.00}  & 71.76 / 49.37   & 77.85 / 56.89 & 23.67 / \textcolor{darkred}{02.03}  \\
& IKE                                    & \textcolor{darkgreen}{99.95} / \textcolor{darkgreen}{99.90}                & \textcolor{darkgreen}{94.40} / \textcolor{darkgreen}{91.22}  & 51.42 / 23.43 & 40.75 / \textcolor{darkred}{05.40}  & \textcolor{darkgreen}{99.40} / \textcolor{darkgreen}{98.94}   & 52.23 / 14.66  & 45.05 / 13.69  \\
& MEND                                    & 20.65 / \textcolor{darkred}{00.00}                & 20.40 / \textcolor{darkred}{00.00} & \textcolor{darkgreen}{96.45} / \textcolor{darkgreen}{89.87}  & \textcolor{darkred}{07.06} / \textcolor{darkred}{00.00}  & 47.04 / \textcolor{darkred}{00.00}   & \textcolor{darkgreen}{90.13} / 70.11 & 22.62 / \textcolor{darkred}{00.00}  \\
& KN                                      & \textcolor{darkred}{08.40} / \textcolor{darkred}{00.00}                 & 10.55 / \textcolor{darkred}{00.00}  & 45.10 / \textcolor{darkred}{04.44}  & \textcolor{darkred}{05.88} / \textcolor{darkred}{00.00}  & 12.19 / \textcolor{darkred}{00.00}  & 37.47 / \textcolor{darkred}{03.95} & 14.02 / \textcolor{darkred}{00.00}  \\
& ROME                                  & 24.88 / \textcolor{darkred}{08.29}                & 20.17 / \textcolor{darkred}{02.51}  & \textcolor{darkgreen}{93.75} / \textcolor{darkgreen}{82.45} & \textcolor{darkred}{07.06} / \textcolor{darkred}{00.00}  & 60.44 / 12.83    & \textcolor{darkgreen}{94.75} / \textcolor{darkgreen}{83.70}  & 24.75 / \textcolor{darkred}{02.12}  \\
& MEMIT                                    & 25.84 /  \textcolor{darkred}{09.60}               & 20.41 / \textcolor{darkred}{02.12}   & \textcolor{darkgreen}{98.67} / \textcolor{darkgreen}{95.76} & \textcolor{darkred}{07.29} / \textcolor{darkred}{00.00}  & 64.16 / 13.31  & \textcolor{darkgreen}{96.75} / \textcolor{darkgreen}{89.49}  & 26.10 / \textcolor{darkred}{02.31}  \\ \bottomrule[1pt]
\end{tabular}            
}
\caption{Experimental results on the \texttt{Chinese-LLaMA-Plus-7B} and \texttt{Chinese-LLaMA-2-7B} backbones in terms of F1 / EM. \textcolor{darkred}{Grey} denotes the score is less than 10.0, while \textcolor{darkgreen}{green} indicates the score is more than 80.0. ``Editing language'' denotes the model is edited by the samples of which language.}
\label{table:chinese_llama_results}
\end{table*}

\noindent \textbf{Cross-Lingual Knowledge Editing.} Given a multi-lingual language model $p_{m\theta}$ and an edit descriptor in a source language $\langle x^{s}_e, y^{s}_e \rangle$, the goal of cross-lingual knowledge editing is to create an edited model $p'_{m\theta}$ satisfy the following requirements:
\begin{equation}
p'_{m\theta} (x^s) = 
\begin{cases}
y^s_e & x^s \in \mathcal{X}^s_e \\
p_{m\theta} (x^s) & x^s \notin \mathcal{X}^s_e 
\end{cases}
\end{equation}
\begin{equation}
p'_{m\theta} (x^t) = 
\begin{cases}
I^t(y^s_e) & x^t \in I^t(\mathcal{X}^s_e) \\
p_{m\theta} (x^t) & x^t \notin I^t(\mathcal{X}^s_e) 
\end{cases}
\end{equation}
where $x^s$ and $x^t$ denote the input text in the source language $s$ and a different target language $t$, respectively.
$\mathcal{X}^s_e$ indicates the edit scope in the source language.
$I^t(\cdot)$ transforms the input text from its source language into the target language $t$ with the same meaning, \emph{i.e.}, translation.
Therefore, in addition to learning edited knowledge in the source language, the model $p'_{m\theta}$ should also reflect consistent behaviors when querying in a different language.
The cross-lingual knowledge editing also needs to satisfy the four properties, \emph{i.e.}, reliability, generality, locality and portability.
Different from the monolingual scenario, all test samples (except reliability samples) in the cross-lingual scenario are in both the source and the target languages, respectively.
For example, an English edited model will be evaluated by a Chinese generality sample to indicate its cross-lingual generality.

\section{Experiments}

\begin{table}[t]
\centering
\setlength{\belowcaptionskip}{-10pt}
\resizebox{0.47\textwidth}{!}
{
\begin{tabular}{lcc}
\toprule[1pt]
                      & C-Eval & MMLU \\ \midrule[1pt]
GPT-4                 & 68.7   & 86.4 \\
GPT-3.5-turbo         & 54.4   & 70.0 \\
\texttt{Baichuan2-7B}           & 54.0   & 54.2 \\
\texttt{Baichuan-7B}           & 42.8   & 42.3 \\
\texttt{Chinese-LLaMA-2-7B}    & 34.4   & 36.8 \\
\texttt{Chinese-LLaMA-Plus-7B} & 25.5   & 31.8 \\ \bottomrule[1pt]
\end{tabular}
}
\caption{Chinese and English capability of the LLMs used in our experiments.}
\label{table:backbone_base_ability}
\end{table}

\subsection{Experimental Setup}

\noindent \textbf{Metrics.} To evaluate the edited model in terms of reliability, generality, locality and portability, different questions, which pair with the golden answers, are input to the edited model.
Thus, we follow previous QA studies~\cite{rajpurkar-etal-2016-squad,yang-etal-2018-hotpotqa} and adapt exact match (EM) and F1 as two evaluation metrics: (1) EM measures the percentage of predictions that match the golden answers exactly. (2) F1 measures the average overlap between the prediction and the golden answer. We treat the prediction and ground truth as bags of tokens, and compute their F1.

\begin{table*}[t]
\centering
\resizebox{0.95\textwidth}{!}
{
\begin{tabular}{llccccccc}
\toprule[1pt]
\multicolumn{1}{c}{Editing} & \multicolumn{1}{c}{\multirow{2}{*}{Method}} & \multirow{2}{*}{Reliability} & \multicolumn{3}{c}{English Evaluation} & \multicolumn{3}{c}{Chinese Evaluation}    \\
\multicolumn{1}{c}{Language}    &                    &                              & Generality               & Locality              & Portability            & Generality               & Locality              & Portability             \\ \midrule[1pt]
\multicolumn{9}{c}{\texttt{Baichuan-7B}} \\ \midrule[1pt]
\multicolumn{1}{c}{\multirow{5}{*}{English}} & FT                                      & 33.33 / 13.11                & 27.09 / \textcolor{darkred}{07.43}  & \textcolor{darkgreen}{91.71} / \textcolor{darkgreen}{83.12}   & \textcolor{darkred}{09.21} / \textcolor{darkred}{00.19}  & 20.79 / \textcolor{darkred}{00.19}  & \textcolor{darkgreen}{87.36} / 64.71 & 30.77 / \textcolor{darkred}{00.10}  \\
& IKE  & \textcolor{darkgreen}{100.0} / \textcolor{darkgreen}{100.0} & \textcolor{darkgreen}{99.72} / \textcolor{darkgreen}{99.61}  & 66.87 / 48.02  & 69.79 / 50.72  & \textcolor{darkgreen}{99.25} / \textcolor{darkgreen}{98.65}  & 47.96 / 15.72   & 44.58 / 14.66 \\
& KN                                    & 10.77 / \textcolor{darkred}{00.00}                & 10.32 / \textcolor{darkred}{00.00}   & 71.28 / 55.74 & \textcolor{darkred}{08.96} / \textcolor{darkred}{00.19}  & 19.69 / \textcolor{darkred}{00.00}   & \textcolor{darkgreen}{93.32} / 80.71  & 31.74 / \textcolor{darkred}{00.00}  \\
& ROME                                    & 68.97 / 52.36                & 60.45 / 42.53     & \textcolor{darkgreen}{98.31} / \textcolor{darkgreen}{96.43}  & \textcolor{darkred}{09.65} / \textcolor{darkred}{00.29}  & 24.45 / \textcolor{darkred}{01.45}  & \textcolor{darkgreen}{98.71} / \textcolor{darkgreen}{95.85} & 31.61 / \textcolor{darkred}{00.29}  \\
& MEMIT                                 & 71.20 / 54.97                & 66.47 / 49.66  & \textcolor{darkgreen}{98.60} / \textcolor{darkgreen}{96.72} & \textcolor{darkred}{09.43} / \textcolor{darkred}{00.10} & 26.19 / \textcolor{darkred}{02.51}   & \textcolor{darkgreen}{98.82} / \textcolor{darkgreen}{95.56}   & 30.53 / \textcolor{darkred}{00.29}  \\ \hline
\multicolumn{1}{c}{\multirow{5}{*}{Chinese}} & FT                             & 13.08 / \textcolor{darkred}{01.45}                & 13.39 / \textcolor{darkred}{00.58}   & \textcolor{darkgreen}{95.18} / \textcolor{darkgreen}{90.26}  & \textcolor{darkred}{09.28} / \textcolor{darkred}{00.29}  & 28.71 / \textcolor{darkred}{04.34}  & 53.83 / 16.88  & 27.76 / \textcolor{darkred}{00.29}  \\
& IKE  & \textcolor{darkgreen}{100.0} / \textcolor{darkgreen}{100.0} & \textcolor{darkgreen}{98.20} / \textcolor{darkgreen}{97.01}  & 70.28 / 51.40  & 69.82 / 51.11  & \textcolor{darkgreen}{100.0} / \textcolor{darkgreen}{100.0}  & 46.92 / 14.95 & 48.91 / 19.38 \\
& KN                 & 10.22 / \textcolor{darkred}{00.00}                & 10.49 / \textcolor{darkred}{00.00}    & 73.43 / 58.24  & \textcolor{darkred}{09.04} / \textcolor{darkred}{00.29} & 19.52 / \textcolor{darkred}{00.00}   & \textcolor{darkgreen}{84.62} / 59.98  & 31.64 / \textcolor{darkred}{00.00}  \\
& ROME                & 24.04 / \textcolor{darkred}{06.36}                & 16.05 / \textcolor{darkred}{01.93}    & \textcolor{darkgreen}{98.06} / \textcolor{darkgreen}{95.66}  & \textcolor{darkred}{09.40} / \textcolor{darkred}{00.29}  & 68.74 / 12.63  & \textcolor{darkgreen}{97.96} / \textcolor{darkgreen}{93.15}  & 27.98 / \textcolor{darkred}{00.68}  \\
& MEMIT      & 23.95 / \textcolor{darkred}{06.27}                & 19.11 / \textcolor{darkred}{05.59}     & \textcolor{darkgreen}{98.47} / \textcolor{darkgreen}{96.53}  & \textcolor{darkred}{09.05} / \textcolor{darkred}{00.19}  & 72.29 / 14.75    & \textcolor{darkgreen}{96.87} / \textcolor{darkgreen}{90.55} & 24.49 / \textcolor{darkred}{00.48}  \\ \midrule[1pt]
\multicolumn{9}{c}{\texttt{Baichuan2-7B}} \\ \midrule[1pt]
\multicolumn{1}{c}{\multirow{5}{*}{English}} & FT  & 33.43 / \textcolor{darkred}{00.48} & 32.25 / \textcolor{darkred}{00.00}  & \textcolor{darkgreen}{90.47} / 78.50  & 27.28 / \textcolor{darkred}{01.74} & 24.76 / \textcolor{darkred}{03.47} & \textcolor{darkgreen}{80.79} / 61.43  & 22.64 / \textcolor{darkred}{00.29} \\
& IKE  & 77.76 / 70.40 & 77.71 / 70.30  & 71.18 / 51.01  & 58.11 / 37.51 & \textcolor{darkgreen}{97.00} / \textcolor{darkgreen}{95.85} & 71.97 / 47.44  & 65.61 / 42.43 \\
& KN  & 10.06 / \textcolor{darkred}{00.00} & \textcolor{darkred}{09.66} / \textcolor{darkred}{00.00}  & \textcolor{darkgreen}{96.22} / \textcolor{darkgreen}{93.83} & 31.62 / \textcolor{darkred}{00.29} & 19.77 / \textcolor{darkred}{00.00}  & \textcolor{darkgreen}{95.51} /  \textcolor{darkgreen}{90.55}  & 25.21 / \textcolor{darkred}{00.58} \\
& ROME  & \textcolor{darkgreen}{88.33} / \textcolor{darkgreen}{81.97} & 73.22 / 59.88  & \textcolor{darkgreen}{95.96} / \textcolor{darkgreen}{90.26} & 31.10 / \textcolor{darkred}{02.31} & 29.82 / \textcolor{darkred}{07.52}  & \textcolor{darkgreen}{96.36} / \textcolor{darkgreen}{89.20}  & 26.00 / \textcolor{darkred}{01.25}  \\
& MEMIT  & \textcolor{darkgreen}{89.34} / \textcolor{darkgreen}{83.32} & \textcolor{darkgreen}{82.54} / 73.19  &  \textcolor{darkgreen}{98.61} / \textcolor{darkgreen}{96.53}  & 30.57 / \textcolor{darkred}{01.74} & 32.11 / 11.09  & \textcolor{darkgreen}{98.31} / \textcolor{darkgreen}{95.08} & 24.59 / \textcolor{darkred}{00.58} \\ \hline
\multicolumn{1}{c}{\multirow{5}{*}{Chinese}} & FT  & 11.45 / \textcolor{darkred}{00.19} & 13.29 / \textcolor{darkred}{00.00}  & \textcolor{darkgreen}{92.46} / \textcolor{darkgreen}{83.03} & 30.11 / \textcolor{darkred}{00.39} & 34.04 / \textcolor{darkred}{07.04}  & 60.57 / 29.80  & 26.78 / \textcolor{darkred}{02.31} \\
& IKE  & \textcolor{darkgreen}{97.08} / \textcolor{darkgreen}{96.05} & 77.36 / 69.72 & 71.48 / 51.88 & 58.58 / 37.99  & \textcolor{darkgreen}{97.00} / \textcolor{darkgreen}{95.95}  & 73.64 / 50.63 & 66.64 / 43.68 \\
& KN  & \textcolor{darkred}{08.92} / \textcolor{darkred}{00.10} & \textcolor{darkred}{08.95} / \textcolor{darkred}{00.10}  & \textcolor{darkgreen}{84.11} / \textcolor{darkgreen}{81.39} & 27.39 / \textcolor{darkred}{00.29} & 18.30 / \textcolor{darkred}{00.10}  & \textcolor{darkgreen}{84.58} / 78.01  & 22.78 / \textcolor{darkred}{00.58}  \\
& ROME  & 35.55 / 19.38 & 17.12 / \textcolor{darkred}{01.06}  & \textcolor{darkgreen}{95.32} / \textcolor{darkgreen}{90.07}  & 31.85 / \textcolor{darkred}{00.48}  & \textcolor{darkgreen}{90.38} / \textcolor{darkgreen}{83.70}  & \textcolor{darkgreen}{95.00} / \textcolor{darkgreen}{85.54} & 28.53 / \textcolor{darkred}{00.96} \\
& MEMIT  & 35.29 / 19.00 & 17.37 / \textcolor{darkred}{01.35}  & \textcolor{darkgreen}{98.74} / \textcolor{darkgreen}{96.24}  & 31.45 / \textcolor{darkred}{00.87}  & \textcolor{darkgreen}{93.45} / \textcolor{darkgreen}{88.52}  & \textcolor{darkgreen}{97.13} / \textcolor{darkgreen}{92.09} & 28.16 / \textcolor{darkred}{00.77}  \\ \bottomrule[1pt]
\end{tabular}
}
\caption{Experimental results on the \texttt{Baichuan-7B} and \texttt{Baichuan2-7B} backbones in terms of F1 / EM.}
\label{table:baichuan_results}
\end{table*}

\vspace{0.5ex}
\noindent \textbf{Baselines.} Following~\citet{Yao2023EditingLL,wang2023easyedit}, we adopted 7 methods as baselines: (1) Directly fine-tuning (\textbf{FT}) the language models with $L_\infty$ constraint;
(2) \textbf{SERAC}~\cite{mitchell2022memory} utilizes a scope classifier to evaluate whether new input is close to the stored editing examples, and further influences the model behaviors based on the retrieved editing examples;
(3) \textbf{IKE}~\cite{zheng2023can} uses context-edit facts to guide the model in generating edited facts;
(4) \textbf{MEND}~\cite{mitchell2022fast} transforms the gradient of fine-tuned language models by employing a low-rank decomposition of gradients;
(5) \textbf{KN}~\cite{dai-etal-2022-knowledge} specifies a key-value pair in the FFN matrix that embodies the knowledge and then proceeds to update the corresponding parameters;
(6) \textbf{ROME}~\cite{meng2022locating} leverages causal mediation analysis to locate the edit area, and updates the whole parameters in the FFN matrix;
(7) \textbf{MEMIT}~\cite{meng2023mass} directly updates LLMs with many memories, and thus, facilitating thousands of edits to be executed simultaneously.

\vspace{0.5ex}
\noindent \textbf{Backbones.} Considering the English and Chinese abilities, we adopt the following four LLMs in the experiments: (1) \texttt{Chinese-LLaMA-Plus-7B}\footnote{\url{https://github.com/ymcui/Chinese-LLaMA-Alpaca}} is created based on LLaMA-7B~\cite{touvron2023llama} with Chinese vocabulary extension and continual pre-training.
(2) In the similar way, \texttt{Chinese-LLaMA-2-7B}\footnote{\url{https://github.com/ymcui/Chinese-LLaMA-Alpaca-2}} is created based on LLaMA-2-7B~\cite{touvron2023llama2}.
(3) \texttt{Baichuan-7B}\footnote{\url{https://github.com/baichuan-inc/Baichuan-7B/}} and (4) \texttt{BaiChuan2-7B}\footnote{\url{https://github.com/baichuan-inc/Baichuan2}} are two LLMs that support both English and Chinese.
Table~\ref{table:backbone_base_ability} lists the above LLMs' performance on C-Eval~\cite{huang2023c} and MMLU~\cite{hendrycks2021measuring} to show their Chinese and English abilities, respectively.
\texttt{Baichuan2-7B} performs the best among the four backbones in both two evaluation benchmark datasets.

\vspace{0.5ex}
\noindent \textbf{Implementation Details.}
All experiments are conducted on a single NVIDIA A800 GPU (80G).
The implementation of all baselines is employed by EasyEdit~\cite{wang2023easyedit} with the default settings.
The hyper-parameters of each method can be found in the corresponding GitHub repository.\footnote{\url{https://github.com/zjunlp/EasyEdit/tree/main/hparams}}

\begin{figure*}[t]
\centering
\subfigure[Chinese-LLaMA (En)]{
  \includegraphics[width=0.22\linewidth]{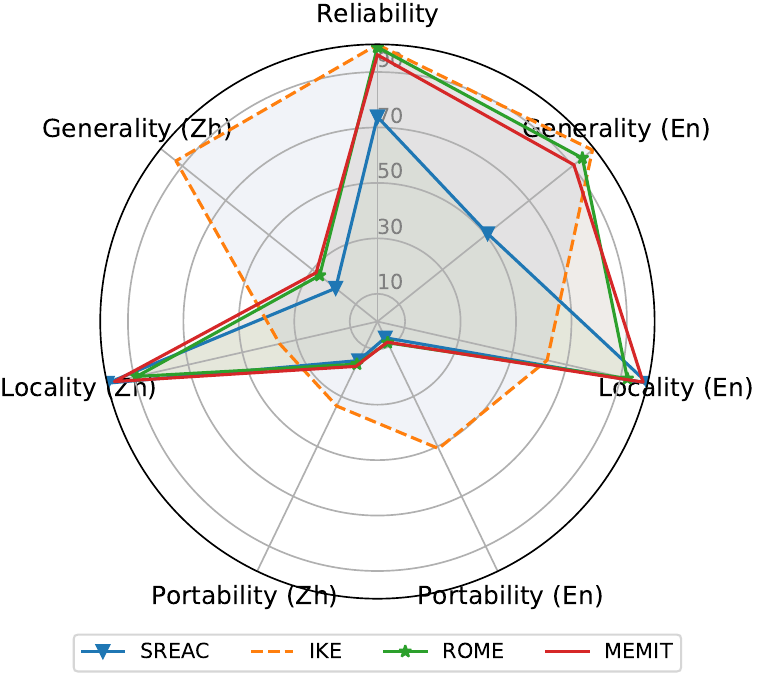}
}
\subfigure[Chinese-LLaMA (Zh)]{
  \includegraphics[width=0.22\linewidth]{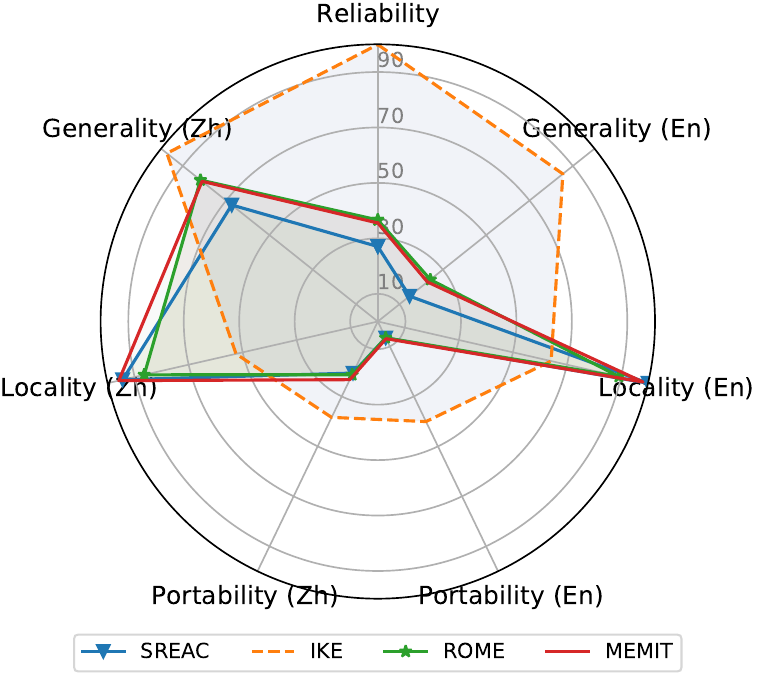}
}
\subfigure[Chinese-LLaMA-2 (En)]{
  \includegraphics[width=0.22\linewidth]{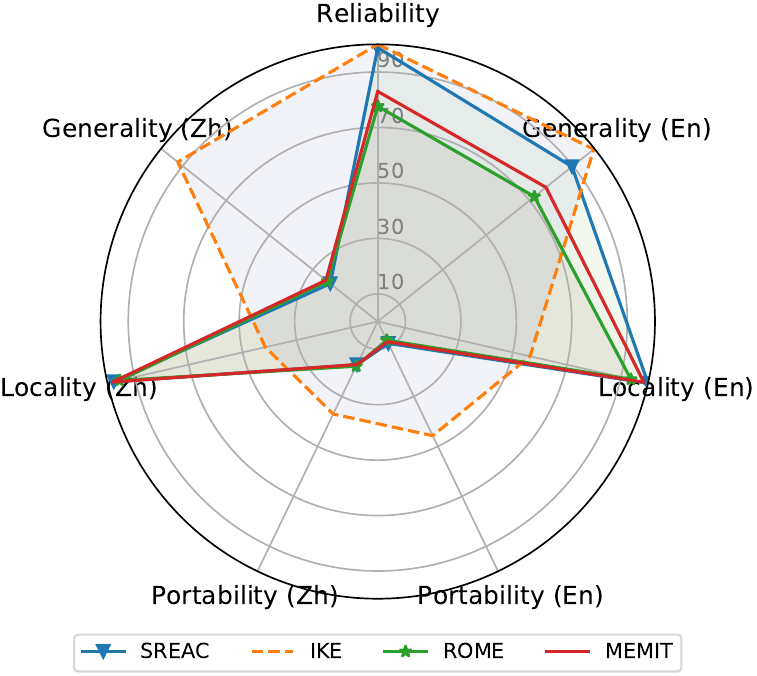}
}
\subfigure[Chinese-LLaMA-2 (Zh)]{
  \includegraphics[width=0.22\linewidth]{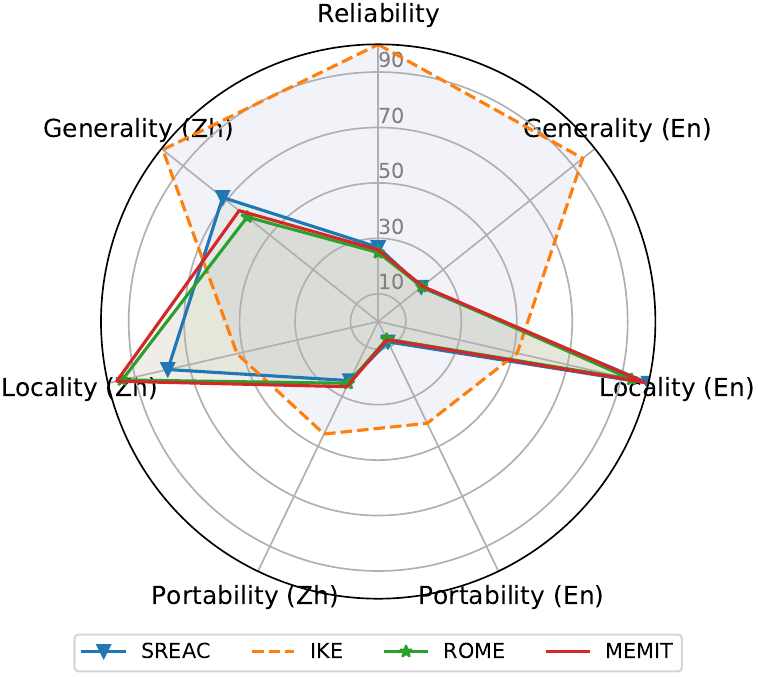}
}
\subfigure[Baichuan (En)]{
  \includegraphics[width=0.22\linewidth]{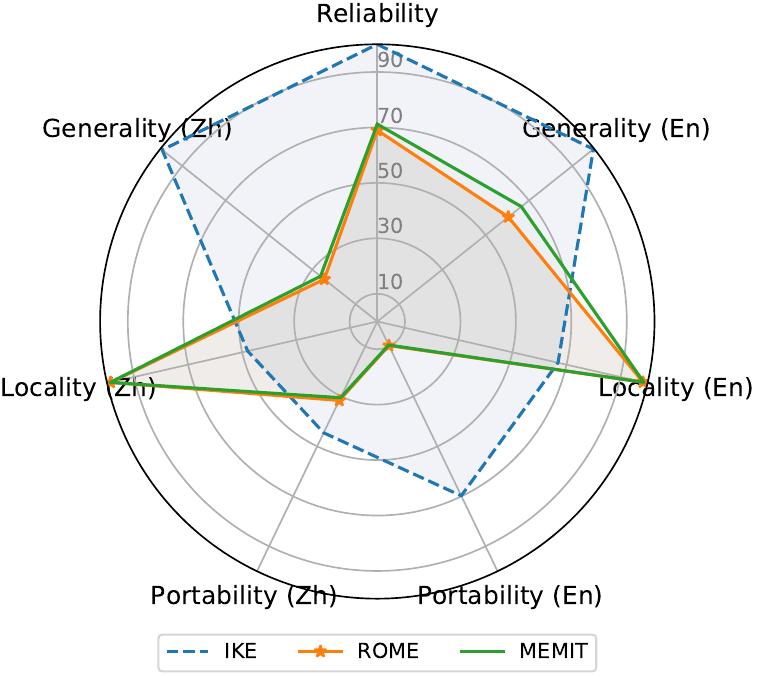}
}
\subfigure[Baichuan (Zh)]{
  \includegraphics[width=0.22\linewidth]{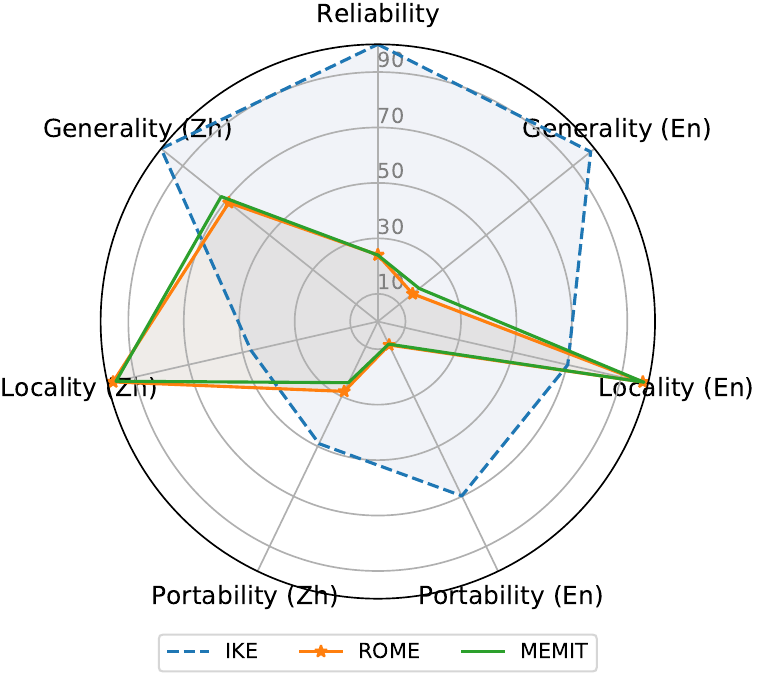}
}
\subfigure[Baichuan2 (En)]{
  \includegraphics[width=0.22\linewidth]{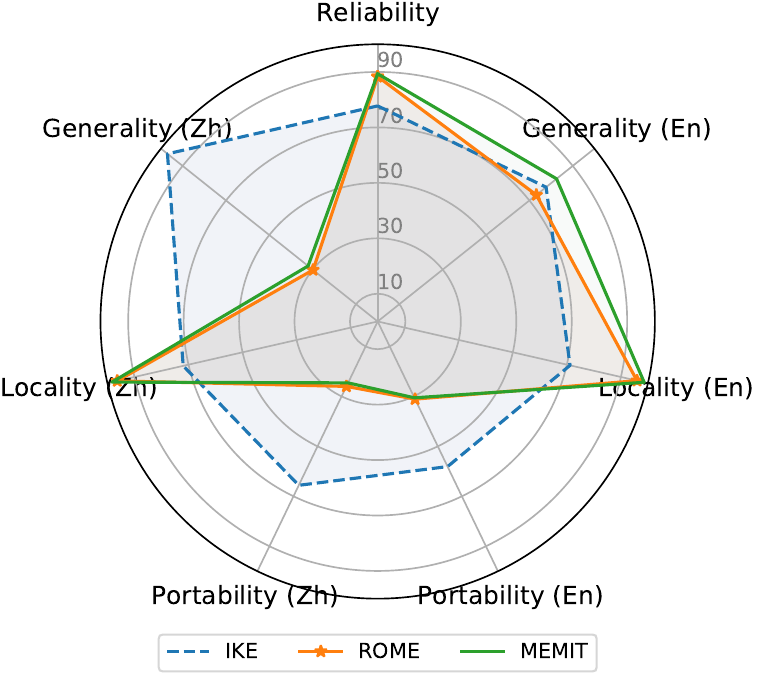}
}
\subfigure[Baichuan2 (Zh)]{
  \includegraphics[width=0.22\linewidth]{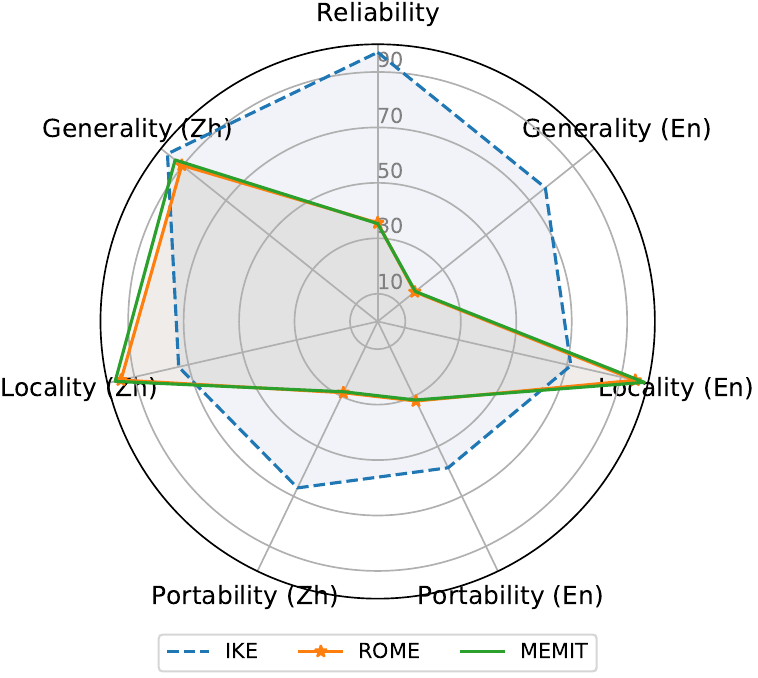}
}
\caption{Radar chart of knowledge editing performance when editing different LLMs with different languages. The language identifiers (\emph{i.e.}, En and Zh) after the name of LLMs indicate the model is edited by the samples of which language, while those after the name of properties indicate the language of the corresponding testing samples.}
\label{fig:radar_chart}
\end{figure*}

\subsection{Results \& Analyses}

Table~\ref{table:chinese_llama_results} shows the experimental results on Chinese-LLaMA-Plus-7B and Chinese-LLaMA-2-7B.

\vspace{0.5ex}
\noindent \textbf{Monolingual Analysis.} Compared with the other three properties, portability is more challenging for knowledge editing methods to achieve. As we can see, IKE achieves the best performance in terms of portability among all baselines.
However, the best EM score in portability is still less than 60.0, showing it is non-trivial to absorb the edited knowledge for most editing methods.
As for reliability which directly evaluates the model performance on the edited knowledge, we find that FT and KN obtain limited performance in this property, thus failing to edit knowledge in LLMs.
For example, KN (En) and KN (Zh) only achieve 4.63 and 3.09 F1 reliability with the Chinese-LLaMA-Plus-7B backbone, while the counterparts of FT (En) and FT (Zh) are 20.46 and 9.54.
Given the above analyses, we next compare the cross-lingual knowledge editing performance on SERAC, IKE, MEND, ROME and MEMIT methods.

\vspace{0.5ex}
\noindent \textbf{Inconsistent Behaviors in Reliability.} When using different languages to edit LLMs, there might be performance gaps in terms of reliability. For example, SERAC (En) achieves 73.84 F1 while SERAC (Zh) only achieves 27.05 F1 on Chinese-LLaMA-Plus-7B. A similar situation could also be found in MEND, ROME and MEMIT.
When using English samples to edit LLMs via these four methods, the achieved reliability is significantly higher than using the Chinese samples.
This is because the language modeling ability of different languages might be different in a single integrated multi-lingual LLM.
Many LLMs show their strong English ability perhaps due to the high-quality English data dominating the pre-training corpora~\cite{touvron2023llama,touvron2023llama2}.
The language modeling gaps of different languages might influence the efficiency of knowledge editing in different languages.
Moreover, we find that when editing LLMs via IKE using different languages, the achieved reliability scores are similar.
For example, IKE (En) and IKE (Zh) achieve the same F1 score (99.90) in terms of reliability on Chinese-LLaMA-Plus-7B.
This indicates that not all knowledge editing methods are sensitive to the choice of editing languages.
Given the strong in-context learning ability of LLMs, IKE can efficiently edit them with demonstration samples in different languages.

\vspace{0.5ex}
\noindent \textbf{Inconsistent Behaviors in Generality.} It is intuitive that when using one language to edit LLMs, the generality in this language is significantly higher than in others. For example, SERAC (En) achieves 50.86 F1 in English generality but only performs 19.26 F1 in Chinese generality (with the Chinese-LLaMA-Plus-7B backbone).
In contrast, SERAC (Zh) achieves better Chinese generality than English (67.41 F1 vs. 14.67 F1).
Almost all knowledge editing methods have this phenomenon.
This finding also indicates that the cross-lingual performance of knowledge editing is still limited.
It is hard for existing knowledge editing methods to transfer the edited knowledge from one language to others in multi-lingual LLMs, and reflect consistent behaviors when querying with different languages.

\begin{figure*}[t]
\centering
\subfigure{
  \includegraphics[width=0.28\linewidth]{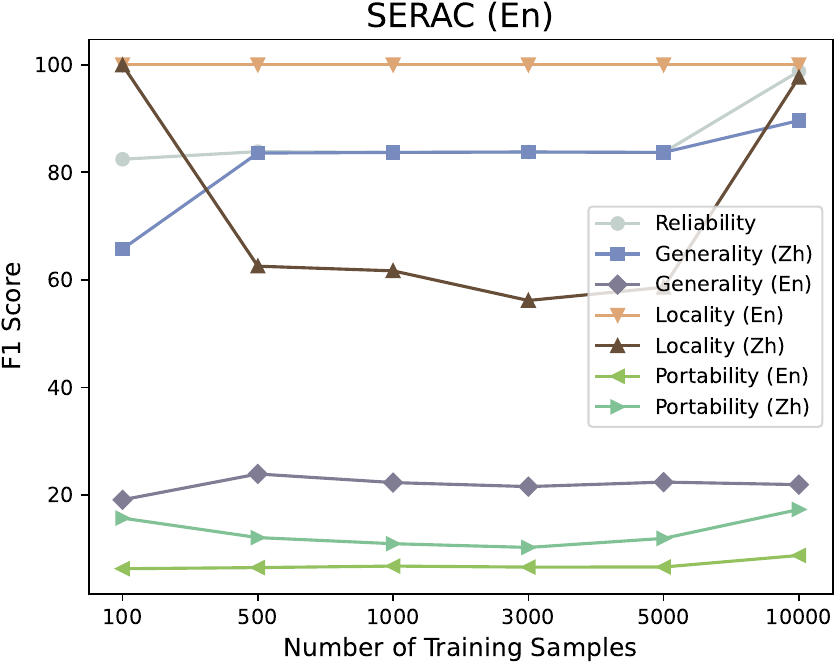}
}
\subfigure{
  \includegraphics[width=0.28\linewidth]{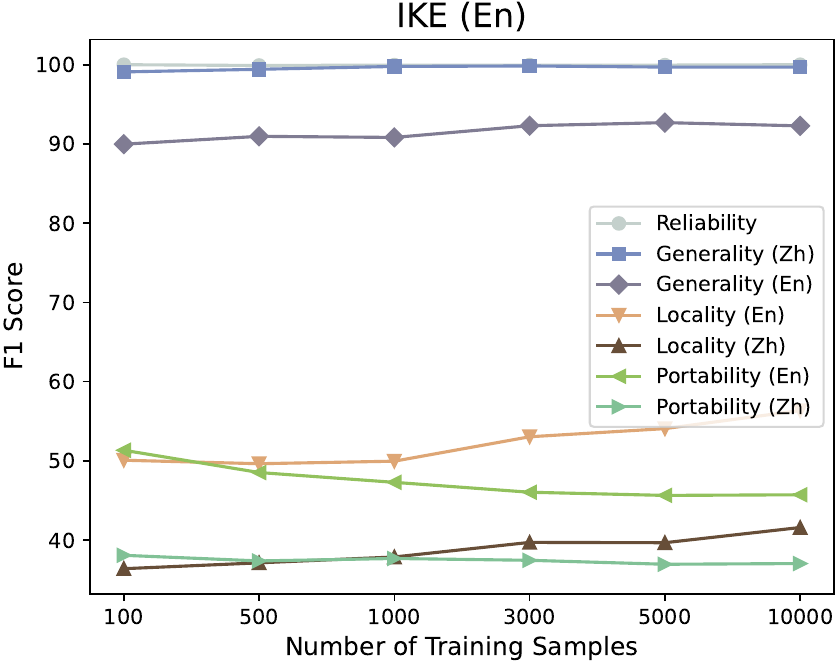}
}
\subfigure{
  \includegraphics[width=0.28\linewidth]{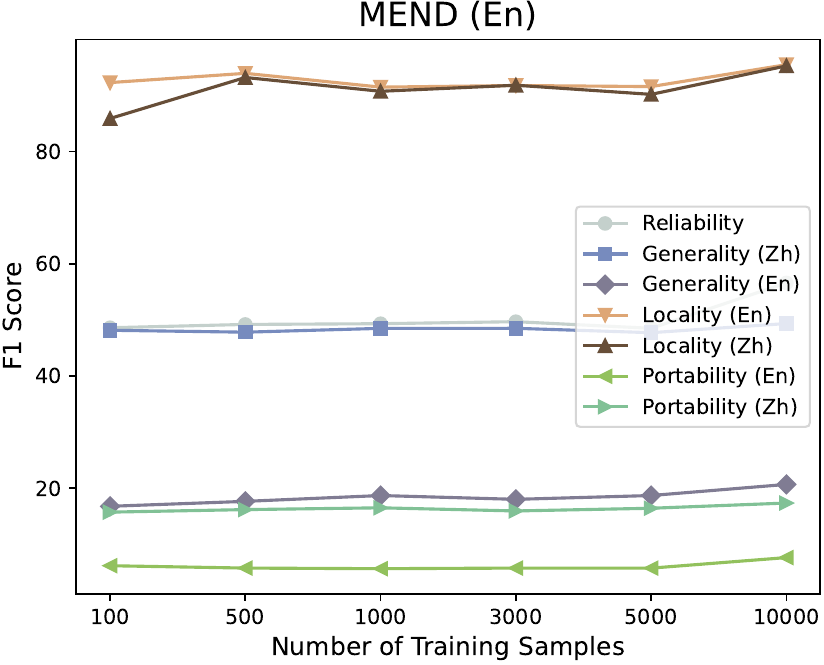}
}
\subfigure{
  \includegraphics[width=0.28\linewidth]{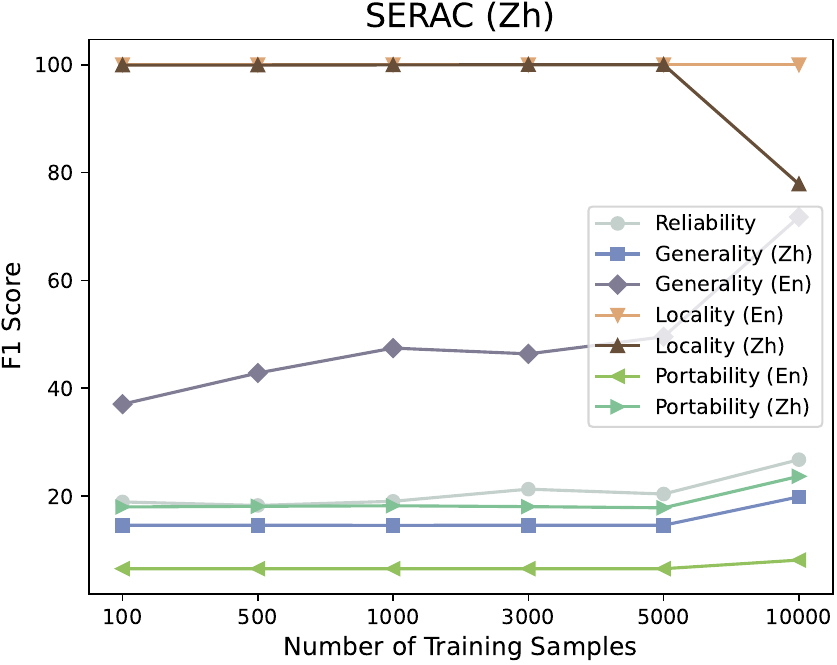}
}
\subfigure{
  \includegraphics[width=0.28\linewidth]{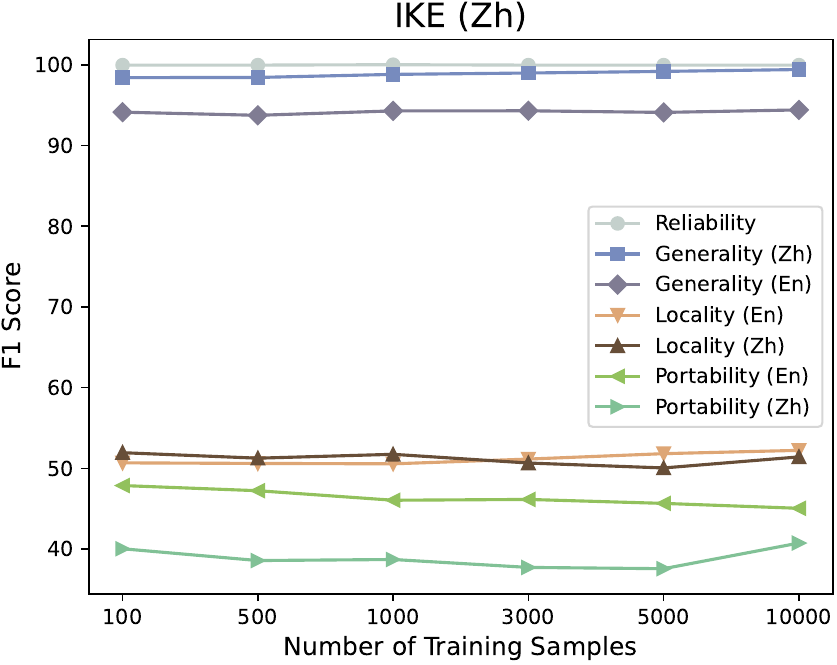}
}
\subfigure{
  \includegraphics[width=0.28\linewidth]{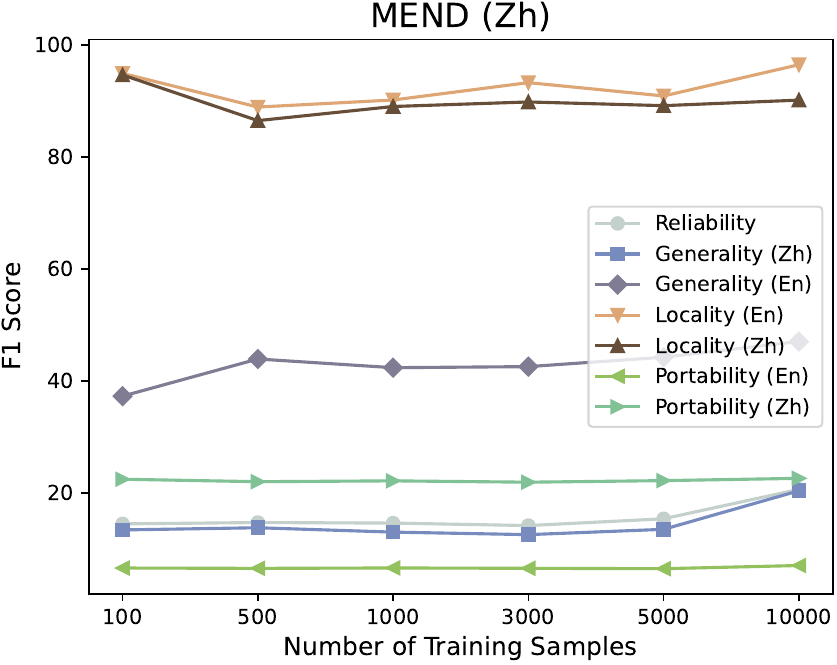}
}
\caption{F1 scores of the edited Chinese-LLaMA-2-7B using different numbers of training samples.}
\label{fig:training_scale}
\end{figure*}

\vspace{0.5ex}
\noindent \textbf{Cross-Lingual Influence on Locality.} When editing LLMs in a source language, the locality in other languages could also be influenced.
When editing Chinese-LLaMA-Plus-7B, MEND (En) achieves 88.96 F1 and 91.75 F1 in English and Chinese locality scores, while the counterparts in MEND (Zh) are 89.19 and 78.17.
Ideally, when editing LLMs in a source language, its performance on irrelevant target-language samples should remain unchanged.
Previous work typically only studies locality in the same language and neglects the cross-lingual locality.
We also find that though IKE works well in terms of reliability and generality, its locality is generally less than that of SERAC, ROME or MEMIT.
The low-level locality makes its usefulness need to be carefully verified in real applications.

\vspace{0.5ex}
\noindent \textbf{Limited Portability in both Languages.} 
As shown in Figure~\ref{fig:radar_chart}, when editing multi-lingual LLMs in English or Chinese, their portability performance in both languages is extremely limited compared with other properties.
This finding indicates that most existing knowledge editing methods only memorize the superficial changes in wording rather than absorbing the edited knowledge.
This phenomenon shows that sharing knowledge across different languages is tricky.
As a result, the currently edited LLMs reflect inconsistent behaviors on the edited knowledge in different languages.

\vspace{0.5ex}
\noindent \textbf{Knowledge Editing Performance on Baichuan.}
Table~\ref{table:baichuan_results} shows the knowledge editing performance on Baichuan-7B and Baichuan2-7B. The results show similar situations to those of Chinese-LLaMA-Plus-7B and Chinese-LLaMA-2-7B, verifying the generality of the previous phenomena and our analyses.

\subsection{The Influence of Training Scale}

Due to the high cost of translating all ZsRE training samples (163K), we randomly select and translate 10K samples via gpt-3.5-turbo (Section~\ref{subsec:3.1}).
We further conduct experiments to investigate the model performance when the training set is limited.
Specifically, we randomly choose 100, 500, 1K, 3K and 5K training samples to conduct experiments.
Among all baselines, SERAC, IKE and MEND are three knowledge editing methods that need additional training.
Thus, we use different numbers of training samples to edit LLMs via these methods, and evaluate their performance in terms of reliability, generality, locality and portability.
Figure~\ref{fig:training_scale} shows the results using Chinese-LLaMA-2-7B as an example backbone.
As we can see, the reliability, generality and portability of SERAC typically increase with the number of training samples, especially SERAC (Zh).
The English locality of SERAC is stable while the Chinese locality is sensitive to the training data.
As for IKE and MEND, their performances are relatively stable in terms of all metrics.
We also find that the portability of IKE and MEND may decrease or remain the same as the number of training samples increases.
Therefore, simply adding more training samples in these two methods cannot increase the ability of the edited models to absorb and reason the edited knowledge. Future work could explore more effective methods or design more reasonable training paradigms to let LLMs go beyond memorizing the superficial changes in wording.

\section{Conclusion}
In this paper, we first explore the cross-lingual effect of knowledge editing in large language models. To achieve that, we automatically construct Bi-ZsRE dataset by translating the previous ZsRE dataset from English to Chinese.
Based on Bi-ZsRE, we conduct experiments on various knowledge editing methods, and study the cross-lingual effect from English to Chinese and vice versa.
Our results indicate that it is still hard for existing knowledge editing methods to transfer the edited knowledge from one language to another in a multi-lingual LLM.
We also analyze the inconsistent behaviors of the edited models and discuss their specific challenges to provide a deeper understanding of the cross-lingual effect in knowledge editing.

\section*{Ethical Considerations}

In this section, we consider the potential ethical issues of our work. In this paper, we propose the Bi-ZsRE dataset which is collected based on the publicly-available datasets, \emph{i.e.}, ZsRE~\cite{levy-etal-2017-zero} and portability QA pairs provided by~\citet{Yao2023EditingLL}. Therefore, Bi-ZsRE might involve the same biases and toxic behaviors exhibited by these datasets.
Besides, we obtain our Bi-ZsRE dataset by translating these datasets, and their corresponding license is the MIT License which is granted to copy, distribute and modify the contents.

During manually correcting the results of machine translation, the salary for each annotator is determined by the average time of annotation and local labor compensation standard.

\section*{Limitation}

While we show the cross-lingual effect in knowledge editing, there are some limitations worth considering in future work:
(1) Bi-ZsRE only involves English and Chinese, and future work could
extend our Bi-ZsRE to more languages and give more comprehensive analyses w.r.t different language families.
(2) The backbones used in our experiments are several LLMs with 7B parameters. Future work can extend our analyses to other LLMs with more parameters (\emph{e.g.}, 13B and 70B).

\section*{Acknowledgement}
This work is partially supported by the National Natural Science Foundation of China (Grant No. 62206056).
We thank anonymous reviewers for their constructive suggestions and comments.

\bibliography{references}

% \appendix

% \section{Example Appendix}
% \label{sec:appendix}

% This is an appendix.

\end{document}